\documentclass[conference]{IEEEtran}
\IEEEoverridecommandlockouts
\usepackage{cite}
\usepackage{amsmath,amssymb,amsfonts}
\usepackage{algorithmic}
\usepackage{graphicx}
\usepackage{textcomp}
\usepackage{latexsym}
\usepackage{amssymb}
\usepackage{amsmath}
\usepackage{amsthm}
\usepackage{bbm}
\usepackage{booktabs}
\usepackage{multirow}
\usepackage{booktabs}
\usepackage{enumitem}
\usepackage{graphicx}
\usepackage{color}
\usepackage{xcolor}
\usepackage{hyperref}

\def\BibTeX{{\rm B\kern-.05em{\sc i\kern-.025em b}\kern-.08em
    T\kern-.1667em\lower.7ex\hbox{E}\kern-.125emX}}
\begin{document}

\title{Zero-shot LLM-guided Counterfactual Generation: A Case Study on NLP Model Evaluation}

\author{\IEEEauthorblockN{Amrita Bhattacharjee}
\IEEEauthorblockA{\textit{School of Computing and AI} \\
\textit{Arizona State University}\\
Tempe, AZ, USA \\
abhatt43@asu.edu}
\and
\IEEEauthorblockN{Raha Moraffah}
\IEEEauthorblockA{\textit{Department of Computer Science} \\
\textit{Worcester Polytechnic Institute}\\
Worcester, MA, USA \\
rmoraffah@wpi.edu
}
\and
\IEEEauthorblockN{Joshua Garland}
\IEEEauthorblockA{\textit{Global Security Initiative} \\
\textit{Arizona State University}\\
Tempe, AZ, USA \\
Joshua.Garland@asu.edu}
\and
\IEEEauthorblockN{Huan Liu}
\IEEEauthorblockA{\textit{School of Computing and AI} \\
\textit{Arizona State University}\\
Tempe, AZ, USA \\
huanliu@asu.edu}

}

\maketitle

\begin{abstract}
With the development and proliferation of large, complex, black-box models for solving many natural language processing (NLP) tasks, there is also an increasing necessity of methods to stress-test these models and provide some degree of interpretability or explainability. While counterfactual examples are useful in this regard, automated generation of counterfactuals is a data and resource intensive process. such methods depend on models such as pre-trained language models that are then fine-tuned on auxiliary, often task-specific datasets, that may be infeasible to build in practice, especially for new tasks and data domains. Therefore, in this work we explore the possibility of leveraging large language models (LLMs) for zero-shot counterfactual generation in order to stress-test NLP models.
We propose a structured pipeline to facilitate this generation, and we hypothesize that the instruction-following and textual understanding capabilities of recent LLMs can be effectively leveraged for generating high quality counterfactuals in a zero-shot manner, without requiring any training or fine-tuning. 
Through comprehensive experiments on a variety of propreitary and open-source LLMs, along with various downstream tasks in NLP, we explore the efficacy of LLMs as zero-shot counterfactual generators in evaluating and explaining black-box NLP models. 


\end{abstract}


\section{Introduction}

Over the last couple of decades, machine learning and natural language processing (NLP) systems have developed massively, especially in terms of the complexity and scale of the models used in different downstream tasks. For example, for most NLP tasks, such as tasks in the GLUE~\cite{wang2018glue} or SuperGLUE~\cite{wang2019superglue} benchmarks, the state-of-the-art performance is achieved by large, black-box models such as pre-trained language models (PLM)~\cite{liu2021towards}. Effective use and deployment of such models, especially in high-stakes areas, require careful evaluation, validation and stress-testing. Furthermore, models should also be explainable or interpretable, i.e., decisions made by such black-box models should ideally be accompanied by how and/or why the model reached that decision~\cite{molnar2020interpretable}. While such endeavors are still challenging in the context of black-box models, in this regard, \textit{counterfactual examples} have been used to perform evaluation, explanation, robustness testing and even improvement of NLP models~\cite{wu2021polyjuice,madaan2021generate,bhattacharjee2024towards}. For example, the following two sentences - \textit{s1}: \textit{This movie is \textcolor{blue}{brilliant!}}, \textit{s2}: \textit{This movie is \textcolor{red}{boring.}} are counterfactual examples for the input sentence \textit{This movie is} \textcolor{blue}{great}. Such minimally perturbed variations of the input text can be used in a variety of settings to evaluate models, to understand whether a model is able to focus on the task-specific features in the input text in order to classify the input text correctly. 

\begin{figure}
    \centering
    \includegraphics[width=0.7\columnwidth]{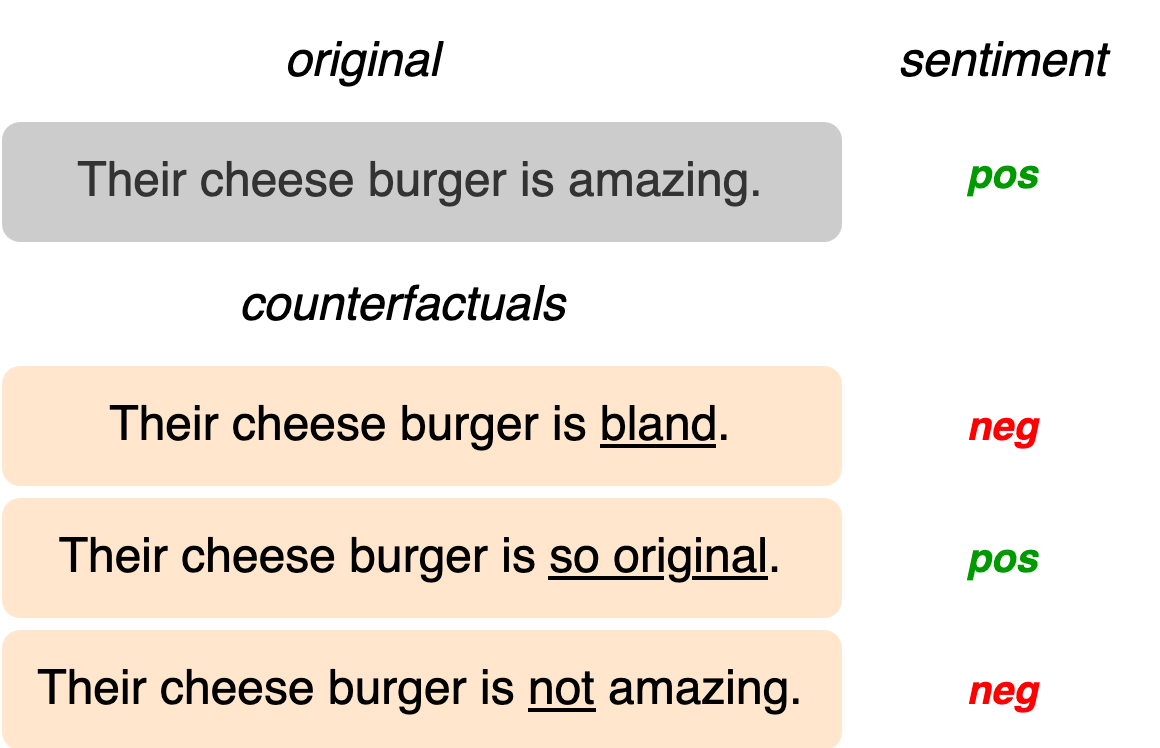}
    \caption{Examples of an input sentence and its corresponding counterfactual examples with same or opposite label.}
    \label{fig:enter-label}
\end{figure}

While several previous works have investigated the applicability of human expert annotators to design such counterfactual examples~\cite{gardner2020evaluating,qin2019counterfactual}, this is not scalable in practice, thereby motivating the exploration of automated counterfactual generation methods. Automated counterfactual generation methods such as ~\cite{wu2021polyjuice,madaan2021generate} use pre-trained language models, or mask-filling, or models trained via control codes for the generation. 
For example, training a conditional generation model in Polyjuice~\cite{wu2021polyjuice} requires sentence-pair dataset for each control code (such as: negation, quantifier, shuffle, lexical, etc.). Similar methods requiring large amounts of training and/or task-specific data are used in other automated counterfactual generation methods. However, having access to such task-specific training datasets may be infeasible in practice, especially for newly emerging data domains and tasks. Therefore, we are interested in investigating: \textit{Is there a way to simplify the counterfactual generation process and perform the generation without any auxiliary data?} 

In order to explore this, in this work, we address a new problem and application setting:\textit{ zero-shot counterfactual generation for evaluating and explaining NLP models}. We tackle this problem by using the power of recent state-of-the-art instruction-tuned large language models (LLMs). While recent work has started exploring the effectiveness of using LLMs for generating counterfactuals, these works use additional guidance, such as, in the form of few-shot exemplars for in-context learning. Given that gold-standard exemplar samples can be hard to come across for many tasks, alongside LLM context length issues restricting the number of in-context examples, we explore the possibility of using LLMs in a \textit{zero-shot} manner for generating counterfactuals in order to evaluate and explain black-box text classifiers. Given that recent LLMs are trained on massive amounts of text data, followed by subsequent supervised fine-tuning and alignment steps, empirical evidence suggests that such LLMs can be used as pseudo-oracles or general-purpose solvers especially in NLP  tasks~\cite{bubeck2023sparks}. As an extension, we propose the paradigm of using LLMs as zero-shot counterfactual generators for stress-testing text classifier models. To further this exploration, we propose a pipeline that leverages recent LLMs in order to generate plausible, human-interpretable counterfactual examples in a completely zero-shot manner. Our proposed pipeline requires only the input text along with either the ground truth label or the predicted label from the black-box classifier (depending on the use-case) and uses a structured, hard-prompting method to use off-the-shelf LLMs for generating the counterfactuals, without any fine-tuning or training with additional data. We envision that automating the task of counterfactual generation via a carefully designed pipeline that leverages LLMs can help to reduce costs and make NLP model development, evaluation and explanation more streamlined and efficient. 
We use our proposed pipeline to generate counterfactuals in order to explore their effectiveness in (1) explaining and, (2) evaluating NLP models for a variety of downstream tasks.
Our results demonstrate that, when used in our pipeline, LLMs may be effectively used to generate effective zero-shot counterfactuals that can be used for stress-testing text classifiers. 

To the best of our knowledge, this is the first piece of work to tackle the problem of zero-shot counterfactual generation to evaluate and explain text classifiers.  
Overall our contributions in this paper are as follows:

\begin{enumerate}
    \item We explore a \textbf{\underline{F}}ramework for \textbf{\underline{I}}nstructed \textbf{\underline{Z}}ero-shot Counterfactual  Generation with \textbf{\underline{L}}anguag\textbf{\underline{E}} Models, which we refer to as \texttt{FIZLE} for brevity \footnote{All code, prompts, supplementary materials, etc. are available at \url{https://github.com/AmritaBh/zero-shot-llm-counterfactual}.}.
    \item We investigate and evaluate \texttt{FIZLE} for two important use-cases: \textit{explaining} and \textit{evaluating} black-box text classification models.
    \item Through experiments on three benchmark datasets, several open-source and proprietary LLMs, we investigate the effectiveness of the proposed pipeline compared to recent baselines and discuss implications for future work in this direction.
\end{enumerate}

\begin{figure}
    \centering
    \includegraphics[width=0.75\columnwidth]{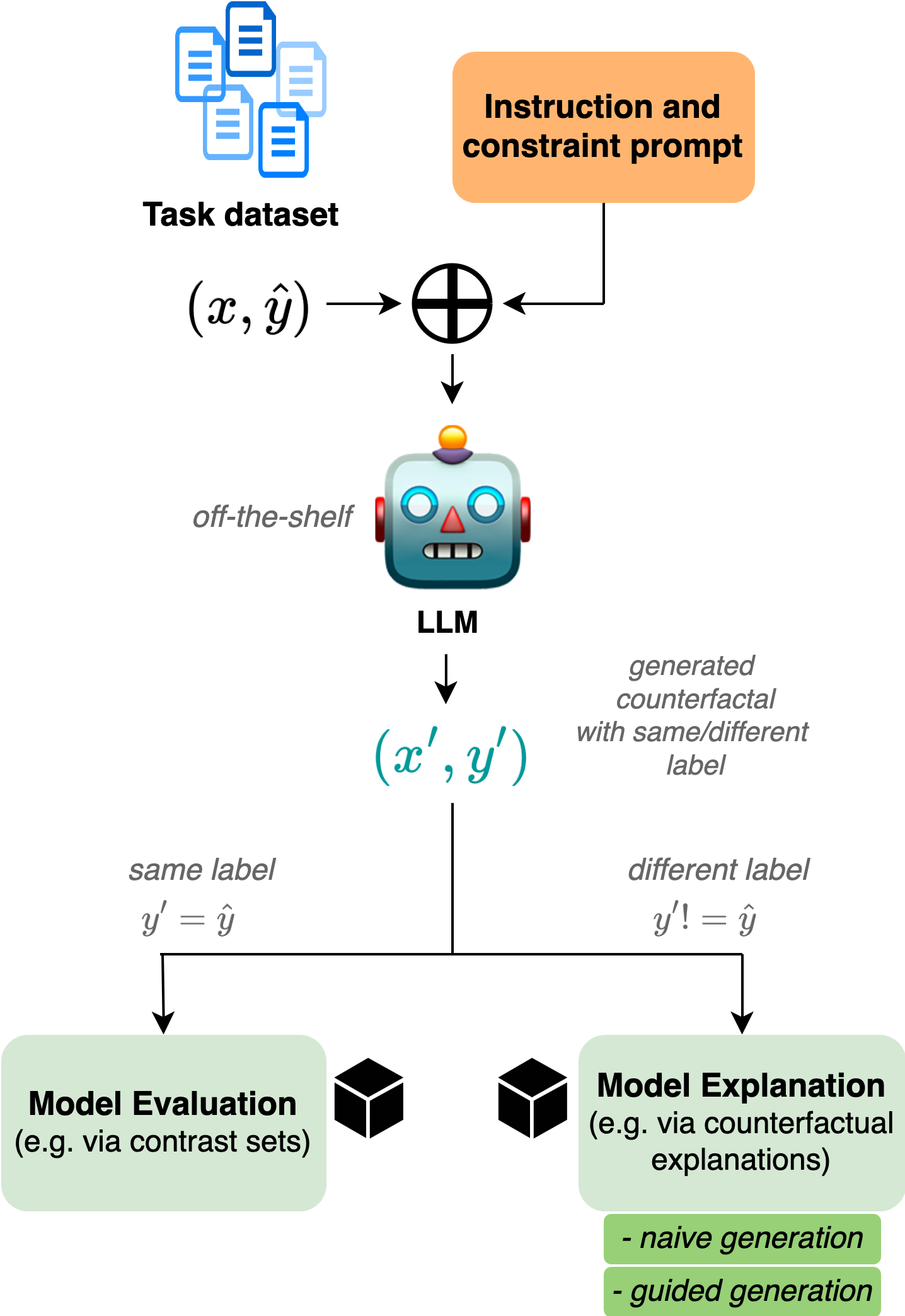}
    \caption{Our proposed \texttt{FIZLE} pipeline for zero-shot LLM-guided counterfactual generation for evaluation and explanation of black-box text classifiers.}
    \label{fig:framework}
\end{figure}


\section{Background and Related Works}
\label{sec:background}


In this section we describe some preliminary concepts along with relevant related works.

\paragraph{Counterfactual Examples} According to most definitions in literature~\cite{molnar2020interpretable}, counterfactuals in text are minimally edited versions of an original text that can flip the label of a classifier\footnote{although, in this paper, we often use the phrase ``counterfactual with same label as input" which effectively refers to a semantically similar re-write of the input text, i.e, similar to having undergone a label-preserving data augmentation step.}. Counterfactuals are typically similar to the input instance, and vary from it in a small number of features. Counterfactual examples may be used to stress-test trained models, provide explanations in the form of counterfactual explanations, and also for model improvement via training with counterfactual examples and counterfactually augmented data. While several efforts have been made in the manual creation of counterfactuals~\cite{gardner2020evaluating,qin2019counterfactual}, this method does not scale up and is therefore infeasible in practice for most use-cases. Automated methods for counterfactual generation are therefore more prevalent. Such methods often use language models trained on some control codes for conditional text generation in order to generate plausible and diverse counterfactuals~\cite{madaan2021generate,wu2021polyjuice}. In the context of text classification, which is the scope of this paper, counterfactual examples can be generated from the input text via token-based substitution methods, masked-language modeling, controlled text generation via control codes~\cite{madaan2022plug}. Authors in ~\cite{robeer2021generating} create realistic counterfactuals via language modeling using a Counterfactual GAN architecture. However, all of these methods use either auxiliary models and/or training data, for example, to capture the style characteristics of different control codes. Some recent work has also explored using large language models in few-shot settings for counterfactual generation~\cite{li2024prompting}. Unlike previous work, in this paper, we focus on \textit{zero-shot counterfactual generation for stress-testing text classifiers}. 


\paragraph{Large Language Models and Applications in NLP} Large Language Models (LLMs) are usually transformer-based models capable of generating human-like text. Recent examples of large language models include the GPT family of models from OpenAI~\cite{radford2019language,brown2020language,openai2023gpt}, the Llama family of models from Meta~\cite{touvron2023llama,touvron2023llama2,dubey2024llama}, Gemini from Google ~\cite{team2023gemini}, etc. A general training recipe for training LLMs include an unsupervised pre-training step, where the model is trained using a huge corpora of text, typically comprising of text from the internet~\cite{penedo2023refinedweb,gao2020pile}, followed by one or more supervised fine-tuning steps, such as instruction-tuning~\cite{ouyang2022training,chung2024scaling}. More recent state-of-the-art LLMs such as GPT-3.5 or GPT-4 are further fine-tuned via reinforcement learning with human feedback (RLHF)~\cite{christiano2017deep}, in order to `align' such models more with human preferences and values. During the fine-tuning and RLHF stages LLMs learn to follow instructions for specific tasks and respond in a helpful manner. Instruction-tuning essentially fine-tunes the model on massive datasets of (instruction, response) pairs, whereby LLMs \textit{learn} to follow instructions in a prompt in order to perform tasks. The vast amount of training, both via the pre-training and the instruction-tuning stages, enables LLMs to perform complex tasks~\cite{bubeck2023sparks}, perform in-context learning~\cite{dong2024survey}, etc.
Recent advancements in LLMs have sparked simultaneous exploration and research into the applicability of these LLMs on a variety of different tasks, such as data labeling ~\cite{he2023annollm,bansal2023large,tan2024large}, text classification ~\cite{sun2023text,bhattacharjee2024fighting}, model explanation~\cite{bhattacharjee2024towards}, etc. We add on to this emerging body of work and explore LLM-guided zero-shot counterfactual generation for NLP model evaluation.

\paragraph{LLMs for Counterfactual Generation} While the use of LLMs for generating counterfactuals is still an emerging direction, some recent works have started exploring the role and effectiveness of LLMs in counterfactual generation~\cite{nguyen2024llms}. While ~\cite{li2024prompting} provides a thorough evaluation on how prompting, model size, task complexity, etc. affect LLM generations of counterfactuals, authors in ~\cite{chen2022disco} look at how large language models can be used to generate counterfactual data for training smaller language models. Authors in ~\cite{bhattacharjee2024towards} explore the use of LLMs for causal explainability using counterfactuals. Unlike prior work in this direction, in this paper we focus on exploring whether LLMs can be used to generate counterfactuals in the zero-shot setting, \textit{specifically} to evaluate and stress test black-box model in a post-training, pre-deployment scenario. 

\section{Zero-shot LLM-guided Counterfactual Generation}

\label{sec:frame}



In this section, we describe our counterfactual generation methodology as shown in Figure \ref{fig:framework}. Following the causal explanation generation procedure in prior work ~\cite{bhattacharjee2024towards}, we use state-of-the-art LLMs in an off-the-shelf manner, without any fine-tuning. We improve upon prior work ~\cite{bhattacharjee2024towards} by expanding and broadening their pipeline into a more general framework that can work for tasks other than causal explanation. Note that in this paper, we formulate and evaluate our pipeline on the broad task of text classification, whereas formulations for other text tasks can be derived in a similar manner. To facilitate this we explain the following components in our framework:

\paragraph{Input Dataset and Other Task-specific Input} The first component in our pipeline takes a task dataset as input and pre-processes it into tuples denoted by $(x_i, \hat{y}_i)$, where $x_i \in X$ denote a text sample in the input dataset $X$, and $\hat{y}_i \in \{0,1,...,k\}$ denote the ground truth label of the corresponding input, in a $k$-class classification problem. Depending on the use-case, we also have black-box access to a text classification model $f(\cdot)$ whereby we get $f(x_i) = y_i$, which is the predicted label. In this case, we also build tuples of the form  $(x_i, y_i)$ for use in the generation step.

\paragraph{LLM as the Counterfactual Generator} We leverage recent state-of-the-art LLMs as the counterfactual generators. Given that these models have been trained on vast amounts of textual data along with extensive instruction tuning, we assume that LLMs can learn to modify and perturb text input to \textit{simulate} how human annotators generate counterfactuals for specific tasks ~\cite{gilardi2023chatgpt}. For this, we use both proprietary models from OpenAI and open-source models from Meta AI, and use carefully crafted instructions and constraint prompts to generate the counterfactuals. Specifically, we use the following proprietary models via the OpenAI API wherever applicable:

\begin{itemize}
    \item \textit{GPT-3.5}\footnote{\url{https://platform.openai.com/docs/models/gpt-3-5-turbo}}: Often referred to as ChatGPT. This is the model that has been explored in a variety of text applications. Specifically we use the \texttt{gpt-3.5-turbo} variant.
    \item \textit{GPT-4}\footnote{\url{https://platform.openai.com/docs/models/gpt-4-turbo-and-gpt-4}}: This is the successor to GPT-3.5 and is known to be more capable. Specifically this model is purported to be able to understand complex instruction better, thereby making it highly suitable to our task of counterfactual generation. We use the \texttt{gpt-4} and \texttt{gpt-4-turbo} versions in our experiments. 
    \item \textit{GPT-4o}\footnote{\url{https://platform.openai.com/docs/models/gpt-4o}}: This is the most recent and flagship model from the OpenAI GPT family of models. Although this model is capable of solving multimodal tasks (including text and visual input and text output), we only use it for text input. According to OpenAI, this model is capable of solving complex, multi-step tasks, thereby making it a suitable candidate for our task. 
    \item \textit{GPT-4o-mini}\footnote{\url{https://platform.openai.com/docs/models/gpt-4o-mini}}: This is a low-cost, lower-latency and possibly smaller\footnote{There is no official information regarding size of GPT-4o vs. GPT-4o-mini} of the previous GPT-4o model. The lightweight nature of this model may be beneficial for researchers looking to use our pipeline for larger datasets.
\end{itemize}

Among the open-source models, we use the following models from Meta's Llama family of models:

\begin{itemize}
    \item \textit{Llama 2 7B}: This is the 7 billion parameter version of Llama 2 model~\cite{touvron2023llama2}. We specifically use the `chat' variant via Huggingface\footnote{\url{https://huggingface.co/meta-llama/Llama-2-7b-chat-hf}}.
    \item \textit{Llama 2 13B}: This is the 13 billion parameter version of Llama 2 model~\cite{touvron2023llama2}. Similar to the previous one, we use the `chat' variant from Huggingface\footnote{\url{https://huggingface.co/meta-llama/Llama-2-13b-chat-hf}}. 
    \item \textit{Llama 3 8B}: This is the 8 billion parameter variant of the more recent Llama 3 model~\cite{dubey2024llama}. We use the `instruct' variant of the model from Huggingface\footnote{\url{https://huggingface.co/meta-llama/Meta-Llama-3-8B-Instruct}},since it is known to be well-suited for following instructions in the prompt.
\end{itemize}


Due to resource constraints, we were unable to use larger variants of Llama 2 and Llama 3 models.

\paragraph{Instruction and Constraint Prompt} To generate the counterfactuals in a zero-shot manner using the chosen LLM, the prompt needs to have informative instructions and constraints to guide the generation. We generate two types of counterfactuals: (1) actual counterfactuals: that is, counterfactuals that have a different label from the original input, according to the definition of counterfactual example. These are used in the counterfactual explanation experiments (see Section \ref{sec:cf-explain}), and (2) counterfactuals with same label as original: These are used in the contrast set experiments (see Section \ref{sec:contrast-set}). For setting (1), we experiment with two variants of the generation process: (i) \textit{naive}: Here the LLM is directly prompted to generate a counterfactual, and (ii) \textit{guided}: Here we use a two-step process - first leveraging the LLM to identify the important input features (i.e., words) that result in the predicted label, and then prompting the same LLM to edit a minimal set of those identified features to generate the counterfactual. 
For ease of extraction of the generated counterfactuals, we also specify an output constraint that allows easy parsing based on a regular expression string match.

\section{Experimental Settings}
\label{sec:exp-set}


We use the pipeline described in the previous section in order to generate counterfactuals and demonstrate these for stress-testing and explaining black-box text classifiers. Specifically our tasks are: (1) Counterfactual explanations for explaining decisions of black-box text classifiers and (2) Evaluating black-box text classification models via contrast sets. To facilitate this, here we go over the datasets used and the general experimental setup for all our generation and evaluation experiments:

\subsection{Datasets}

In this work, we focus on two broad categories of language tasks: text classification and natural language inference (NLI)\footnote{NLI here is also treated as a text classification task where the labels for each input are simply one of \{entailment, neutral, contradiction\}.}. For text classification we use two datasets: IMDB~\cite{maas-EtAl:2011:ACL-HLT2011} for sentiment classification and AG News\footnote{\url{https://huggingface.co/datasets/ag\_news}} for news topic classification. For NLI, we use the SNLI dataset~\cite{maccartney2008modeling, bowman2015large}. This variety of datasets allows us to evaluate the LLM-generated counterfactuals over a variety of label situations from binary to multi-class. 

The \textbf{IMDB} dataset\footnote{\url{https://huggingface.co/datasets/imdb}} consists of a total of 50k highly polar movie reviews from IMDB (Internet Movie Database). Each data instance consists of a text string comprising the review text, and a label, either `negative' or `positive'. The \textbf{AG News} dataset consists of over 120k news articles, belonging to one of four news topics: `world', `sports', `business' and `science/technology'. The \textbf{Stanford Natural Language Inference (SNLI)} dataset\footnote{\url{https://huggingface.co/datasets/snli}} consists of 570k sentence pairs consisting of a premise and a hypothesis. Each premise-hypothesis pair is labeled with one of `entailment', `contradiction' or `neutral' labels. 

\subsection{Experimental Setup}

All experiments on open-source models were performed on two A100 GPUs with a total of 80G memory. For 13B Llama 2 models, we use 4-bit quantization using the optimal `nf4' datatype~\cite{dettmers2023qlora}. For all LLMs, we use top\_p sampling with $p=1$, temperature $t=0.4$ and a repetition penalty of $1.1$. We use PyTorch and fine-tuned models hosted on HuggingFace in both Sections \ref{sec:cf-explain} and \ref{sec:contrast-set}.

\section{Counterfactual Explanations via LLM-generated Counterfactuals}
\label{sec:cf-explain}


Explainability is a major challenge in many NLP applications such as text classification~\cite{ribeiro2016should,atanasova2020diagnostic,camburu2018snli}. Although recent models involving pre-trained transformer-based language models~\cite{vaswani2017attention} have achieved or even exceeded human-level performance on several tasks~\cite{wang2018glue,wang2019superglue}, most of these models are black-box by design and hence are not interpretable. Such models do not offer transparency on \textit{why} it predicted a certain label, or even \textit{what features in the input} resulted in the prediction. The ubiquity of these black-box classifiers necessitates the development of explanation frameworks and techniques that provide some degree of understanding into the decision-making function of the model~\cite{ali2023explainable}. 
Counterfactual explanations~\cite{molnar2020interpretable} give an insight into what \textit{could have been} different in the input to change the output label predicted by the classifier.
Gold-standard counterfactual generation requires human annotators and is also task-specific~\cite{Kaushik2020Learning,khashabi2020more}, therefore making it an extremely expensive and labor-intensive endeavor. Therefore, we investigate whether we use zero-shot LLM-generated counterfactuals as counterfactual explanations for black-box text classifiers. 

\begin{table*}[]
\centering
\resizebox{\textwidth}{!}{%
\begin{tabular}{@{}cl@{}}
\toprule
\textbf{Framework Variant} &
  \multicolumn{1}{c}{\textbf{Prompt Structure}} \\ \midrule

\multirow{2}{*}{\texttt{FIZLE}$_{guided}$} &
  \begin{tabular}[c]{@{}l@{}}\textbf{Step 1}: In the task of  \textcolor{blue}{\textless{}task on task-dataset\textgreater{}}, a trained black-box classifier correctly predicted \\ the label `\textcolor{purple}{\textless{}$y_i$\textgreater{}}' for the following text. Explain why the model predicted the `\textcolor{purple}{\textless{}$y_i$\textgreater{}}' label by\\ identifying the words in the input that caused the label. \\ List ONLY the words as a comma separated list.\textbackslash{}n---\textbackslash{}nText: \textcolor{teal}{\textless{}$x_i$\textgreater{}}\end{tabular} \\ \cmidrule(l){2-2} 
 &
  \begin{tabular}[c]{@{}l@{}}\textbf{Step 2}: Generate a counterfactual explanation for the original text by ONLY changing a minimal set of the \\ words you identified, so that the label changes from `\textcolor{purple}{\textless{}$y_i$\textgreater{}}' to `\textcolor{olive}{\textless{}$y_{cf}$\textgreater{}}'. Use the following definition of \\ `counterfactual explanation': ``A counterfactual explanation reveals what should have been different in an \\ instance to observe a diverse outcome."  Enclose the generated text within \textless{}new\textgreater{} tags.\end{tabular} \\ \midrule
\texttt{FIZLE}$_{naive}$ &
  \begin{tabular}[c]{@{}l@{}}In the task of \textcolor{blue}{\textless{}task on task-dataset\textgreater{}}, a trained black-box classifier correctly predicted the label `\textcolor{purple}{\textless{}$y_i$\textgreater{}}' \\ for the following text. Generate a counterfactual explanation by making minimal changes to the input text, \\ so that the label changes from `\textcolor{purple}{\textless{}$y_i$\textgreater{}}' to `\textcolor{olive}{\textless{}$y_{cf}$\textgreater{}}'. Use the following definition of `counterfactual explanation': \\ ``A counterfactual explanation reveals what should have been different in an instance to observe a diverse \\ outcome." Enclose the generated text within \textless{}new\textgreater{} tags.\textbackslash{}n---\textbackslash{}nText: \textcolor{teal}{\textless{}$x_i$\textgreater{}}.\end{tabular} \\ \bottomrule
\end{tabular}%
}

\label{tab:prompt}
\end{table*}

\subsection{Methodology}

To generate counterfactual explanations for a black-box text classifier $f(\cdot)$ that predicts $f(x_{i}) = y_{i}$, we use the tuple $(x_{i}, y_{i})$ in the generation step, thereby replacing the ground truth label in Figure \ref{fig:framework} by the model-predicted label, since we aim to explain why the model predicted $y_{i}$ for the input sample $x_{i}$. In our experiments, we use a DistilBERT model ~\cite{sanh2019distilbert} , fine-tuned on the specific task dataset as the black-box model we aim to explain. Note that since our counterfactual generation process is model-agnostic, the same procedure can be applied to any black-box classifier in place of DistilBERT. Inspired by prior work~\cite{bhattacharjee2024towards}, we develop and experiment with two variants of \texttt{FIZLE}: (1) \texttt{FIZLE$_{naive}$}: which directly generates the counterfactual explanation, and (2) \texttt{FIZLE$_{guided}$}: which first extracts words that may have been responsible for the predicted label, and then uses those selected words to generate a counterfactual explanation, in a two-step manner. We hypothesize that the two-step generation may result in more effective and better quality counterfactual explanations, due to the additional guidance provided to the LLM, analogous to prior work such as Chain of Thought prompting~\cite{wei2022chain}. We show the prompts used in both the variants in Table \ref{tab:prompt}.

\subsection{Evaluation Metrics}

To evaluate the goodness of the counterfactual explanations generated by our zero-shot LLM-guided pipeline, we use a variety of evaluation metrics following prior work ~\cite{wu2021polyjuice,madaan2021generate,bhattacharjee2024towards}. Ideally, the generated counterfactual explanations should be able to flip the label of the classifier, thereby showcasing what \textit{could have} changed in the input that would flip the label of the classifier. Furthermore, counterfactual explanations should also be \textit{minimally edited} samples of the input text, i.e., they should be as close as possible to the input sample both in the token space and the semantic space. To capture and evaluate these criteria, we use the following metrics:




\paragraph{Label Flip Score} We use Label Flip Score to measure the effectiveness of the generated counterfactual explanations. For each input text $x_i$ in the test split of the dataset, with correctly predicted label $f(x_i) = y_k$, we evaluate the corresponding LLM-generated counterfactual $x_i^{cf}$ using the same black-box classifier $f(\cdot)$ and obtain a label for the counterfactual. For an effective counterfactual, the obtained label should be different from the original label $y_k$. Then Label Flip Score $\%$ (LFS) is computed as:\\
    \begin{equation}
    \label{eq:eq1}
        LFS = \frac{1}{n}\sum_{i=1}^n \mathbbm{1}[f(x_i) \neq f(x_i^{cf})] \times 100
    \end{equation}

where $n$ is the number of samples in the test set and $\mathbbm{1}$ is the identity function.

\paragraph{Textual Similarity} Counterfactual explanations generated by the LLMs should ideally be as `similar' to the original input text as possible. To evaluate this similarity, we use two metrics: similarity of the text embeddings using the Universal Sentence Encoder (USE)~\cite{cer2018universal} in the latent space, and a normalized Levenshtein distance~\cite{levenshtein1966binary} to measure word edits in the token space. The semantic similarity using the embeddings of the original input and the generated counterfactual is computed as the inner product of the original and the counterfactual embeddings, averaged over the test dataset: 

\begin{equation}
\label{eq:eq2}
    sim_{semantic} = \frac{1}{n}\sum_{i=1}^n Enc(x_i)\cdot Enc(x_i^{cf})
\end{equation}

where $Enc(\cdot)$ refers to the Universal Sentence Encoder, $n$ is the number of samples in the test set.

Levenshtein distance~\cite{levenshtein1966binary} between two strings is defined as the minimum number of single character edits that are required to convert one string to another. To measure the distance between the original input text and the generated counterfactual in the token space we use a normalized Levenshtein distance, further averaged over the test dataset. This is computed as:

\begin{equation}
\label{eq:eq3}
    edit\_dist = \frac{1}{n}\sum_{i=1}^n \frac{lev(x_i, x_i^{cf})}{max(|x_i|, |x_i^{cf}|)}
\end{equation}

where $|x_i|$ and $|x_i^{cf}|$ refer to the length of $x_i$ and $x_i^{cf}$ respectively, $lev(\cdot,\cdot)$ refers to the Levenshtein distance, and $n$ is the number of samples in the test set.


\subsection{Baselines}

Similar to other counterfactual generation methods~\cite{madaan2021generate,wu2021polyjuice}, we compare our proposed \texttt{FIZLE} pipeline with three representative baselines from three categories of similar works: (i) BAE~\cite{garg2020bae} is a recent adversarial attack method that uses masked language modeling with BERT to perturb the input text by replacing masked words; (ii) CheckList ~\cite{ribeiro2020beyond} is a method for behavioral testing of NLP models via test cases generated by template-based methods as well as masked language models like RoBERTa; (iii)  Polyjuice ~\cite{wu2021polyjuice} is a recent counterfactual generation method that uses an auxiliary language model (such as GPT-2) to generate diverse counterfactuals. Note that unlike these baselines, our \texttt{FIZLE} pipeline does not require any additional dataset or training, thereby enabling a completely zero-shot generation.

\subsection{Results: Effectiveness of Generated Counterfactual Explanations}
\begin{table*}[]
\centering
\caption{Evaluation results of both variants of our \texttt{FIZLE} framework in comparison to baselines: BAE~\cite{garg2020bae}, CheckList~\cite{ribeiro2020beyond} and Polyjuice~\cite{wu2021polyjuice}. We report the Label Flip Score (LFS), semantic similarity (Sem. Sim) and normalized Levenshtein distance (Edit Dist.). Best LFS scores for each dataset are in \textbf{bold}, second best is \underline{underlined}. }
\resizebox{\textwidth}{!}{%
\begin{tabular}{@{}cccccccccc@{}}
\toprule
\multirow{2}{*}{\textbf{Model}}                                           & \multicolumn{3}{c}{\textbf{IMDB}}       & \multicolumn{3}{c}{\textbf{AG News}}     & \multicolumn{3}{c}{\textbf{SNLI}}       \\ \cmidrule(l){2-4} \cmidrule(l){5-7} \cmidrule(l){8-10}
                                                                 & LFS $\uparrow$  & Sem. Sim. $\uparrow$ & Edit Dist. $\downarrow$ & LFS $\uparrow$   & Sem. Sim. $\uparrow$ & Edit Dist. $\downarrow$ & LFS $\uparrow$  & Sem. Sim. $\uparrow$ & Edit Dist. $\downarrow$\\ \midrule
BAE~\cite{garg2020bae}                                                              & 79.6  & 0.99    & 0.044     & 25.0     & 0.97    & 0.063     & 74.4  & 0.95      & 0.054     \\
CheckList~\cite{ribeiro2020beyond}                                                        & 2.6   & 0.99      & 0.013      & 1.6    & 0.92    & 0.083     & 3.0     & 0.96     & 0.036      \\
Polyjuice~\cite{wu2021polyjuice}                                                        & 96.86 & 0.25    & 0.884     & 72.64 & 0.22     & 0.749     & \textbf{95.8}  & 0.74      & 0.367     \\ \midrule
\begin{tabular}[c]{@{}c@{}}GPT-3.5 \\ $(guided)$\end{tabular}        & 78.52 & 0.91 & 0.126 & 30.55 & 0.95 & 0.084 & 32.47 & 0.89 & 0.102 \\
\begin{tabular}[c]{@{}c@{}}GPT-3.5 \\ $(naive)$\end{tabular} & 59.19 & 0.88      & 0.236      & 49.0     & 0.91      & 0.325      & 56.39 & 0.92      & 0.182      \\ \midrule
\begin{tabular}[c]{@{}c@{}}GPT-4 \\ $(guided)$\end{tabular}         & 97.2  & 0.89      & 0.142      & 82.39  & 0.65      & 0.232      & 73.6  & 0.88      & 0.153      \\
\begin{tabular}[c]{@{}c@{}}GPT-4\\ $(naive)$\end{tabular}        & \underline{99.6}  & 0.87      & 0.226      & \underline{84.39}  & 0.65      & 0.278      & 78.0    & 0.88      & 0.152      \\ \midrule
\begin{tabular}[c]{@{}c@{}}GPT-4o \\ $(guided)$\end{tabular}         &  93.56 &  0.85     &   0.389    &  62.5 &  0.74     & 0.1917      & 66.56  & 0.92      & 0.092      \\
\begin{tabular}[c]{@{}c@{}}GPT-4o\\ $(naive)$\end{tabular}        &  99.57 &  0.72     & 0.56      & \textbf{90.12}  &  0.41   &  0.549    &  \underline{79.39}  & 0.90    & 0.157      \\ \midrule
\begin{tabular}[c]{@{}c@{}}GPT-4o-mini \\ $(guided)$\end{tabular}         & 98.58 &  0.82    & 0.413    & 66.33 &  0.76   &  0.173     & 47.09  & 0.81      &  0.185     \\
\begin{tabular}[c]{@{}c@{}}GPT-4o-mini\\ $(naive)$\end{tabular}        & \textbf{100.0} & 0.63     &  0.642    & 66.6  &  0.59    & 0.485      & 51.4   & 0.83     & 0.282     \\ \midrule \midrule

\begin{tabular}[c]{@{}c@{}}Llama 2 7B \\ $(guided)$\end{tabular}    & 76.64 & 0.66      & 0.546      & 51.11  & 0.77      & 0.244      & 36.82 & 0.74      & 0.304      \\
\begin{tabular}[c]{@{}c@{}}Llama 2 7B \\ $(naive)$\end{tabular}  & 64.7  & 0.59      & 0.68       & 35.25  & 0.70       & 0.492      & 58.33 & 0.62      & 0.577      \\ \midrule
\begin{tabular}[c]{@{}c@{}}Llama 2 13B \\ $(guided)$\end{tabular}   & 51.11 & 0.70       & 0.533      & 51.65  & 0.77      & 0.266      & 50.2  & 0.67      & 0.495      \\
\begin{tabular}[c]{@{}c@{}}Llama 2 13B \\ $(naive)$\end{tabular}      & 66.67 & 0.52 & 0.715 & 37.63 & 0.58 & 0.606 & 59.95 & 0.55 & 0.621 \\ \midrule
\begin{tabular}[c]{@{}c@{}}Llama 3 8B \\ $(guided)$\end{tabular}    & 90.95 & 0.80      & 0.453      & 47.48  & 0.73      &  0.197     & 44.89 &  0.81    & 0.216      \\
\begin{tabular}[c]{@{}c@{}}Llama 3 8B \\ $(naive)$\end{tabular}  &  97.34 & 0.71      & 0.522       &  75.36 &   0.71     & 0.505      & 61.6 & 0.71      & 0.466      \\ \bottomrule
\end{tabular}%
}
\label{tab:cf-explain}
\end{table*}


Following the experimental setup described above, we evaluate whether generated counterfactuals can be used to explain black-box classifiers for the three datasets, as compared to the baselines. We show these quantitative results in Table \ref{tab:cf-explain}.
For each LLM, we evaluate both variants of our framework: \texttt{FIZLE$_{guided}$} and \texttt{FIZLE$_{naive}$}. For effective and good quality counterfactual explanations, ideally we would expect high values of LFS and semantic similarity with low values of edit distance. Overall, we see varied performance of the LLMs and the two variants across the different tasks. Similar to other counterfactual generation works~\cite{madaan2021generate}, we see an obvious trade-off between the Label Flip Score and the semantic similarity. This is intuitive since the more the generated counterfactual deviates from the original input text, higher the chances are for it to be a successful counterfactual for the original input (i.e, it would result in a label flip). 
Among the three baselines, we see CheckList fails completely in generating counterfactual explanations. We see satisfactory performance by BAE, except for the AG News dataset. For Polyjuice, even though the LFS scores are high, the unsatisfactory textual similarity scores imply that the counterfactuals generated are not good quality and deviate from the input text significantly.

Overall we do not see a clear winner between the two variants: \texttt{FIZLE$_{guided}$} outperforms \texttt{FIZLE$_{naive}$}. However, while the LFS scores are better in the \textit{naive} variant over the \textit{guided}, the LLM is unable to preserve the textual similarity in comparison to the \textit{guided} case. This may imply that the additional `guidance' provided by identifying the input words before the counterfactual explanation generation step enables the generation of counterfactuals closer in semantics to the original input. 
GPT-4o and GPT-4o-mini when used in the $naive$ variant of our pipeline, have the best performance for zero-shot counterfactual explanation generation, in terms of LFS, for IMDB and AG News datasets, respectively. 
For natural language inference on the SNLI dataset, we see all LLMs struggle to generate good counterfactual explanations. GPT-4o performs well, possibly owing to its instruction and textual understanding capabilities~\cite{bubeck2023sparks}, but the best performance is by the Polyjuice baseline. This poor performance of LLMs particularly on the SNLI dataset is further evidence towards LLMs struggling with inference and reasoning. This gap in the capabilities of recent LLMs on reasoning tasks has been observed by several recent efforts as well~\cite{rae2021scaling,valmeekam2022large}.

Lastly, we see the open-source models Llama 2 7B and 13B struggle to generate zero-shot counterfactual explanations with small number of edits, thus resulting in very high edit distances. The Llama 2 models struggle to keep the generated counterfactuals semantically similar to the original input, implying they either make too many edits to the input text, or output some unrelated, low-quality text that does not conform to the instructions provided in the prompt. However, the newer Llama 3 8b model outperforms both these models in most settings.




\renewcommand{\arraystretch}{1.3}

\begin{table*}[]
\centering
\caption{Performance of \texttt{FIZLE}-generated counterfactuals as contrast sets. C.s. Acc. refers to accuracy on the generated contrast sets, Original Test Acc. is the test accuracy on the corresponding paired original samples. Sem. Sim. refers to semantic similarity as computed by Eq. \ref{eq:eq2}, Edit. Dist. is the token level distance as computed by Eq. \ref{eq:eq3}, Cons. (\%) is the consistency as computed by Eq. \ref{eq:eq4}.}
\resizebox{\textwidth}{!}{%
\begin{tabular}{@{}cccccccccccccccc@{}}
\toprule
\multirow{2}{*}{\begin{tabular}[c]{@{}c@{}}\textbf{Counterfactual}\\ \textbf{generator}\end{tabular}} &
  \multicolumn{5}{c}{\textbf{IMDB}} &
  \multicolumn{5}{c}{\textbf{SNLI}} &
  \multicolumn{5}{c}{\textbf{AG News}} \\ \cmidrule(l){2-6} \cmidrule(l){7-11} \cmidrule(l){12-16} 
 &
  \multicolumn{1}{c|}{\begin{tabular}[c]{@{}c@{}}Original  \\ Test Acc.\end{tabular}} &
  C.s. Acc. $\downarrow$ &
  \begin{tabular}[c]{@{}c@{}}Edit \\ Dist. $\downarrow$\end{tabular} &
  \begin{tabular}[c]{@{}c@{}}Sem. \\ Sim. $\uparrow$\end{tabular} &
  \multicolumn{1}{c|}{\begin{tabular}[c]{@{}c@{}}Cons.\\  \%$\downarrow$\end{tabular}} &
  \multicolumn{1}{c|}{\begin{tabular}[c]{@{}c@{}}Original \\ Test Acc.\end{tabular}} &
  C.s. Acc. $\downarrow$ &
  \begin{tabular}[c]{@{}c@{}}Edit \\ Dist. $\downarrow$\end{tabular} &
  \begin{tabular}[c]{@{}c@{}}Sem. \\ Sim. $\uparrow$\end{tabular} &
  \multicolumn{1}{c|}{\begin{tabular}[c]{@{}c@{}}Cons.\\ \%$\downarrow$\end{tabular}} &
  \multicolumn{1}{c|}{\begin{tabular}[c]{@{}c@{}}Original\\ Test Acc.\end{tabular}} &
  C.s. Acc. $\downarrow$&
  \begin{tabular}[c]{@{}c@{}}Edit \\ Dist. $\downarrow$\end{tabular} &
  \begin{tabular}[c]{@{}c@{}}Sem. \\ Sim. $\uparrow$\end{tabular} &
  \begin{tabular}[c]{@{}c@{}}Cons.\\ \%$\downarrow$\end{tabular} \\ \midrule
\multicolumn{1}{c|}{Polyjuice~\cite{wu2021polyjuice}} &
  \multicolumn{1}{c|}{94.3} &
  84.9 &
  - &
  - &
  \multicolumn{1}{c|}{76.1} &
  \multicolumn{1}{c|}{86.5} &
  72.3 &
  - &
  - &
  \multicolumn{1}{c|}{56.4} &
  \multicolumn{1}{c|}{-} &
  - &
  - &
  - &
  - \\
\multicolumn{1}{c|}{Expert~\cite{gardner2020evaluating}} &
  \multicolumn{1}{c|}{96.31} &
  84.84 &
  0.136 &
  0.939 &
  \multicolumn{1}{c|}{81.56} &
  \multicolumn{1}{c|}{-} &
  - &
  - &
  - &
  \multicolumn{1}{c|}{-} &
  \multicolumn{1}{c|}{-} &
  - &
  - &
  - &
  - \\ \midrule
\multicolumn{1}{c|}{GPT-3.5} &
  \multicolumn{1}{c|}{} &
  88.82 &
  0.162 &
  0.931 &
  \multicolumn{1}{c|}{85.22} &
  \multicolumn{1}{c|}{} &
  57.42 &
  0.175 &
  0.908 &
  \multicolumn{1}{c|}{53.53} &
  \multicolumn{1}{c|}{} &
  93.42 &
  0.287 &
  0.883 &
  92.07 \\
\multicolumn{1}{c|}{GPT-4} &
  \multicolumn{1}{c|}{\multirow{4}{*}{93.35}} &
  92.65 &
  0.157 &
  0.942 &
  \multicolumn{1}{c|}{90.95} &
  \multicolumn{1}{c|}{\multirow{4}{*}{86.25}} &
  73.01 &
  0.277 &
  0.841 &
  \multicolumn{1}{c|}{68.82} &
  \multicolumn{1}{c|}{\multirow{4}{*}{94.6}} &
  94.0 &
  0.352 &
  0.855 &
  93.0 \\
\multicolumn{1}{c|}{GPT-4o} &
  \multicolumn{1}{c|}{} &
  95.4 &
  0.133 &
  0.953 &
  \multicolumn{1}{c|}{93.2} &
  \multicolumn{1}{c|}{} &
  80.61 &
  0.237 &
  0.882 &
  \multicolumn{1}{c|}{75.15} &
  \multicolumn{1}{c|}{} &
  95.2 &
  0.406 &
  0.829 &
  92.8 \\
\multicolumn{1}{c|}{GPT-4o-mini} &
  \multicolumn{1}{c|}{} &
  94.0 &
  0.394 &
  0.873 &
  \multicolumn{1}{c|}{91.0} &
  \multicolumn{1}{c|}{} &
  74.34 &
  0.343 &
  0.842 &
  \multicolumn{1}{c|}{68.89} &
  \multicolumn{1}{c|}{} &
  93.8 &
  0.452 &
  0.818 &
  92.0 \\
\multicolumn{1}{c|}{Llama 2 7B} &
  \multicolumn{1}{c|}{} &
  87.32 &
  0.559 &
  0.728 &
  \multicolumn{1}{c|}{83.74} &
  \multicolumn{1}{c|}{} &
  63.88 &
  0.382 &
  0.782 &
  \multicolumn{1}{c|}{57.87} &
  \multicolumn{1}{c|}{} &
  92.94 &
  0.438 &
  0.808 &
  91.72 \\
\multicolumn{1}{c|}{Llama 2 13B} &
  \multicolumn{1}{c|}{} &
  84.76 &
  0.580 &
  0.710 &
  \multicolumn{1}{c|}{82.2} &
  \multicolumn{1}{c|}{} &
  48.6 &
  0.476 &
  0.738 &
  \multicolumn{1}{c|}{43.21} &
  \multicolumn{1}{c|}{} &
  93.5 &
  0.427 &
  0.808 &
  92.03 \\ 
\multicolumn{1}{c|}{Llama 3 8B} &
  \multicolumn{1}{c|}{} &
  89.78 &
  0.267 &
  0.861 &
  \multicolumn{1}{c|}{88.03} &
  \multicolumn{1}{c|}{} &
  72.75 &
  0.158 &
  0.923 &
  \multicolumn{1}{c|}{67.51} &
  \multicolumn{1}{c|}{} &
  93.64 &
  0.229 &
  0.874 &
  91.95 \\ \bottomrule
\end{tabular}%
}
\label{tab:cs-res}
\end{table*}

\section{Evaluating Models via LLM-generated Counterfactuals}
\label{sec:contrast-set}


Deep learning models such as text classification models are often trained in a supervised manner using labeled training sets, and then evaluated on a hold-out test set. Such train-test splits of data usually arise from the same corpus that has the same or similar sources and annotation guidelines. Therefore, in essence, standard evaluation using such hold-out test sets measure merely the in-distribution performance of the model, while in reality, the same model may demonstrate sub-par performance on out-of-distribution or in-the-wild test data~\cite{gardner2020evaluating}. To alleviate this issue to some degree, approaches such as evaluating using challenge sets or robustness to label-preserving perturbations, etc. have been explored by the community. 
One specific method of stress-testing such models is via \textit{contrast sets} ~\cite{gardner2020evaluating}. A contrast set $C(x)$ is essentially a sample of points around a data point $x$, that is close to the local ground truth decision boundary. Samples in $C(x)$ may have same or different ground truth label as $x$. In practice, $C(x)$ can be a set of samples that are `close' to $x$, i.e., have minimal edit distance from $x$, yet be `challenging' for a trained model to classify. In the original contrast sets work~\cite{gardner2020evaluating}, the authors advocate for an evaluation paradigm where dataset authors themselves create and release such contrast sets for model evaluation. However, we note that this is highly infeasible in practice, given the cost of expert creation of such challenging data points. Therefore, automated methods for designing such challenging evaluation sets in the form of contrast sets are highly desirable, albeit at the expense of trading off expert insights. One such automated method for developing contrast sets to evaluate models is that of counterfactual examples, as demonstrated by previous work~\cite{wu2021polyjuice}. Motivated by the effectiveness of counterfactuals as contrast sets in prior work~\cite{wu2021polyjuice}, we envision the use of LLM-generated contrast sets as well for the same purpose of model evaluation. Here we describe the methodology for the generation and evaluation of such contrast sets using LLMs in a zero-shot manner.

\subsection{Methodology}

For generating the contrast sets, we use the same LLMs as used in Section \ref{sec:cf-explain}, and prompt each LLM to generate counterfactuals in a zero-shot manner using the input text and ground truth label tuple $(x_{i}, \hat{y}_{i})$. Unlike ~\cite{wu2021polyjuice}, we do not use human annotators to label the generated counterfactuals. Therefore, differing from ~\cite{wu2021polyjuice}, in our evaluation, we only focus on counterfactuals that have the same label as the original input, and use these as contrast sets. We make this choice since the lack of human annotation and lack of step-by-step guidance (such as in \texttt{FIZLE$_{guided}$}) would make it harder to validate whether the edits performed by the LLM are actually label flipping or not.
Instead, we guide the generation process via the instruction in the prompt. We use the following prompt to perform the generation:

\vspace{1.5mm}
\noindent\fbox{%
    \parbox{0.95\columnwidth}{%
        `You are a robustness checker for a machine learning algorithm. In the task of \textcolor{blue}{\textless{}$task_i$\textgreater{}}, the following data sample has the ground truth label \textcolor{blue}{\textless{}$\hat{y}_{i}$\textgreater{}}. Make minimal changes to the data sample to create a more challenging data point while keeping the ground truth label the same. Text: \textcolor{blue}{\textless{}$x_{i}$\textgreater{}}'

    }%
}
\vspace{1.5mm}

where, $task_i$ is the description of the task, such as ``sentiment classification'', $x_{i}$ is the input text, and $\hat{y}_{i}$ is the ground truth label. 

\subsection{Evaluation Metrics}

For evaluating the goodness of the generated counterfactuals as contrast sets, we compare the accuracy of the target model $f(\cdot)$ on both the original test set and the generated contrast set. Following ~\cite{wu2021polyjuice}, we also measure the consistency, 

\begin{equation}
    consistency = \frac{1}{n}\sum_{i=1}^n \mathbbm{1}[f(x_i) = \hat{y}_{i} \wedge f(x_i^{cs}) = \hat{y}^{cs}_i] \times 100
    \label{eq:eq4}
\end{equation}

where $x_i^{cs}$ is the LLM-generated contrast set for the original input $x_i$, $\hat{y}^{cs}_i$ is the ground truth label for the contrast set example and $n$ is the number of test samples. Consistency measures the percentage of times when the model correctly classifies both the original and the contrast set example.
Like the previous set of experiments, we want the generated counterfactuals (or contrast sets) to be as close to the original text input as possible, i.e., the edits should ideally be minimal. Therefore, we capture the textual similarity again in the token space via Equation \ref{eq:eq3} and the latent space via Equation \ref{eq:eq2}.

\subsection{Baselines}

Since there is not much work on contrast sets, we have a limited set of baselines here. We use the original expert-created contrast sets for the IMDB dataset from the original work~\cite{gardner2020evaluating}. This consists of 488 original test data samples, and 488 contrast samples created by the dataset experts. Furthermore, we use Polyjuice-generated counterfactuals~\cite{wu2021polyjuice} as contrast sets for comparison. 

\subsection{Results: Effectiveness of Counterfactuals as Contrast Sets}

We use the same DistilBERT models as in Section \ref{sec:cf-explain} that are fine-tuned for each of the 3 tasks (IMDB, SNLI and AG News). We evaluate each of these 3 models on both the original test set and the counterfactual one (i.e., the contrast sets) and show these results in Table \ref{tab:cs-res}. We obtain the performance values for Polyjuice-generated contrast sets from the original paper ~\cite{wu2021polyjuice}. For the `Expert' baseline, the IMDB contrast sets are created by human experts in ~\cite{gardner2020evaluating}. Unfortunately, there does not exist any expert created contrast set for SNLI and AG News datasets. As evident from the test accuracies on both the original test set and the counterfactual one, we see a consistent decrease in performance on the generated counterfactuals over the original samples. The performance drop for the AG News dataset seems to be the least while interestingly, we see the highest performance drop on contrast sets for the SNLI dataset. Furthermore, we see GPT-3.5 and GPT-4 are able to create contrast sets with high degree of semantic similarity and low edit distance, thus being more desirable over Llama 2 generated contrast sets. Interestingly, we see that GPT-4o and GPT-4o-mini contrast sets often have better performance than the original test set, which could imply that these contrast sets are too `easy', and therefore not functional. Overall, for the other LLMs, the drop in accuracy and the consistency values seem analogous to similar results in literature (average drop in classification accuracy of around 6.8\% according to ~\cite{wu2021polyjuice}) that use human-generated contrast sets for evaluation ~\cite{gardner2020evaluating,wu2021polyjuice}. Evaluating models with such LLM-generated contrast sets may thereby allow the model developer to investigate what type of samples the model is failing on, thereby informing choices regarding further robustness training.

While this is promising, we do note the ethical concerns surrounding this: LLM-generated contrast sets may induce pre-existing biases that can propagate further bias and errors through evaluation and subsequent model improvement steps. One hybrid way to effectively use LLM-generated contrast sets is by broadly identifying the failure models of the model via probing the model using the LLM-generated contrast sets, and \textit{then} employing human annotators or data creators to hone in on that specific failure mode to either generate more contrast sets or counterfactually augmented training data to fill the identified gap. Such a combined method would greatly reduce costs while still being effective in terms of model evaluation and development.


\section{Case Study: How has the performance of LLMs evolved over time?}

\begin{figure}
    \centering
    \includegraphics[width=0.7\columnwidth]{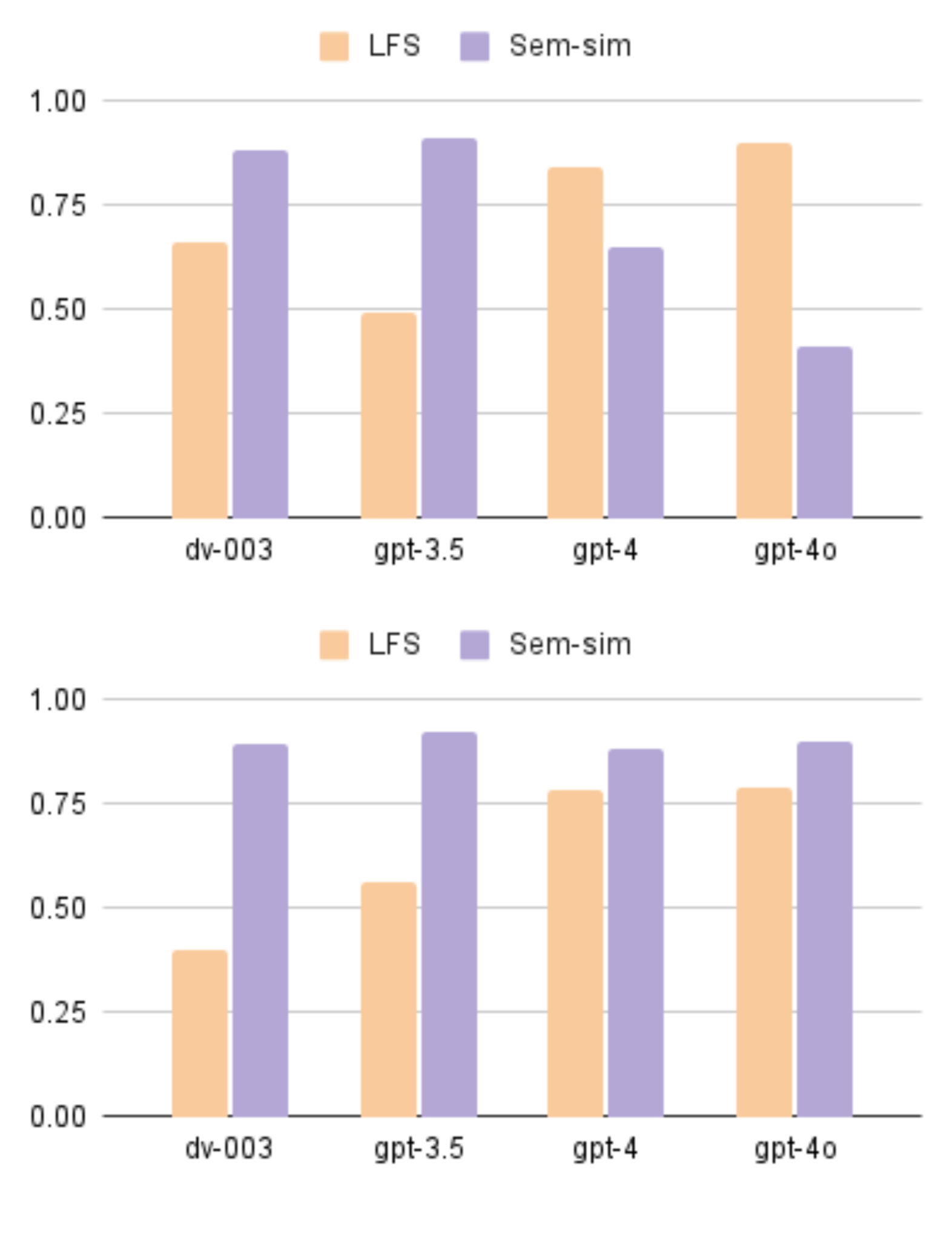}
    \caption{Comparison of LFS and semantic similarity (Sem-sim) for generated counterfactual explanations for AG News (top) and SNLI (bottom). LFS $\%$ scaled to 0-1, higher values for both are better. dv-003 refers to text-davinci-003.}
    \label{fig:gpt-case-study}
\end{figure}

In order to use LLMs for zero-shot counterfactual generation for stress-testing models, one needs to account for the fast evolving landscape of LLM development and release. These models, especially proprietary ones, undergo frequent model updates. Older models are often deprecated, making adoption of such solutions challenging without thorough re-evaluation of pipelines such as ours. To explore the effect of model release and deprecation life cycles, we conduct a small case study to evaluate how performance of this method of generating counterfactuals vary over time. We do this by reporting naive generation results for counterfactual explanations for GPT family of models. Models we use here, from oldest (2022) to newest (2024), are: \texttt{text-davinci-003}\footnote{\url{https://platform.openai.com/docs/deprecations}} $\rightarrow$ \texttt{GPT-3.5} $\rightarrow$ \texttt{GPT-4} $\rightarrow$ \texttt{GPT-4o}.
We show this comparison in Figure \ref{fig:gpt-case-study}. Keeping the expected trade-off between LFS and semantic similarity in mind, we see that for a task like NLI, newer models in the GPT family perform better, achieving high LFS scores while retaining semantic similarity, whereas for AG News, performance is varied: recent models like GPT-4 and GPT-4o achieve higher LFS but at the cost of textual similarity. While a more thorough evaluation is required in order to make broad claims, however, given these tasks and models, recent models like GPT-4o seem to be better at more complex tasks such as reasoning and inference, while a relatively older model like GPT-4 may be sufficient for other types of tasks.

\section{Conclusion \& Future Work}
\label{sec:conclu}


In this paper, we explore the possibility of using LLM-generated zero-shot counterfactuals for stress-testing black-box text classification models. To this end, we propose a pipeline, with two variants, for generation of such counterfactuals in a zero-shot manner. We conduct experiments with a variety of proprietary and open LLMs, and evaluate generated counterfactuals on two broad NLP model development tasks: (1) counterfactual explanation of black-box text classifiers, and (2) evaluation of black-box text classification models via contrast sets. 
Our results are promising and we see benefits to using our proposed \texttt{FIZLE} pipeline across the two use-cases and three downstream tasks. Our findings suggest the effective use of LLM-generated counterfactuals for explaining black-box NLP models, as well as potentially identifying failure models of NLP models via evaluation with contrast sets. We further discuss implications and how hybrid human-and-AI methods may benefit from our exploration, along with looking at how model updates over time can affect the quality and performance of generated counterfactuals.

Future work may investigate modifications to the generation pipeline such as plugging in an additional model, perhaps a small language mode (SLM) to evaluate label of generated contrast sets. More effort can also be put into validating the faithfulness of the generated counterfactual explanation and correctness of generated contrast sets. Further exploration could look into categorizing the failure modes of the black-box models, after being evaluated by these generated counterfactuals. This information can then be used to collect or generate more data for robustness training of these models.
Since one of the challenges in our method was to ensure that the generated text is actually a counterfactual, devising ways and human annotations to ensure more conformity of the generated counterfactual explanations to the definition of `counterfactual explanation' is also be an area that can be improved. Finally, apart from these two use-cases of counterfactuals, LLM-generated counterfactuals can also be evaluated in tasks such as model training or improvement, uncovering biases in model predictions, incorporating fairness into model predictions, etc.

\section*{Acknowledgments}

This work is supported by the DARPA SemaFor project (HR001120C0123). The views, opinions and/or findings expressed are those of the authors.

\bibliographystyle{IEEEtran.bst}
\bibliography{references}

\begin{thebibliography}{10}
\providecommand{\url}[1]{#1}
\csname url@samestyle\endcsname
\providecommand{\newblock}{\relax}
\providecommand{\bibinfo}[2]{#2}
\providecommand{\BIBentrySTDinterwordspacing}{\spaceskip=0pt\relax}
\providecommand{\BIBentryALTinterwordstretchfactor}{4}
\providecommand{\BIBentryALTinterwordspacing}{\spaceskip=\fontdimen2\font plus
\BIBentryALTinterwordstretchfactor\fontdimen3\font minus \fontdimen4\font\relax}
\providecommand{\BIBforeignlanguage}[2]{{%
\expandafter\ifx\csname l@#1\endcsname\relax
\typeout{** WARNING: IEEEtran.bst: No hyphenation pattern has been}%
\typeout{** loaded for the language `#1'. Using the pattern for}%
\typeout{** the default language instead.}%
\else
\language=\csname l@#1\endcsname
\fi
#2}}
\providecommand{\BIBdecl}{\relax}
\BIBdecl

\bibitem{wang2018glue}
A.~Wang, A.~Singh, J.~Michael, F.~Hill, O.~Levy, and S.~R. Bowman, ``Glue: A multi-task benchmark and analysis platform for natural language understanding,'' \emph{arXiv preprint arXiv:1804.07461}, 2018.

\bibitem{wang2019superglue}
A.~Wang, Y.~Pruksachatkun, N.~Nangia, A.~Singh, J.~Michael, F.~Hill, O.~Levy, and S.~Bowman, ``Superglue: A stickier benchmark for general-purpose language understanding systems,'' \emph{Advances in neural information processing systems}, vol.~32, 2019.

\bibitem{liu2021towards}
X.~Liu, T.~Sun, J.~He, J.~Wu, L.~Wu, X.~Zhang, H.~Jiang, Z.~Cao, X.~Huang, and X.~Qiu, ``Towards efficient nlp: A standard evaluation and a strong baseline,'' \emph{arXiv preprint arXiv:2110.07038}, 2021.

\bibitem{molnar2020interpretable}
C.~Molnar, \emph{Interpretable machine learning}.\hskip 1em plus 0.5em minus 0.4em\relax Lulu. com, 2020.

\bibitem{wu2021polyjuice}
T.~Wu, M.~T. Ribeiro, J.~Heer, and D.~S. Weld, ``Polyjuice: Generating counterfactuals for explaining, evaluating, and improving models,'' in \emph{Proceedings of the 59th Annual Meeting of the ACL and the 11th IJCNLP (Volume 1: Long Papers)}, 2021, pp. 6707--6723.

\bibitem{madaan2021generate}
N.~Madaan, I.~Padhi, N.~Panwar, and D.~Saha, ``Generate your counterfactuals: Towards controlled counterfactual generation for text,'' in \emph{Proceedings of the AAAI Conference on Artificial Intelligence}, vol.~35, no.~15, 2021, pp. 13\,516--13\,524.

\bibitem{bhattacharjee2024towards}
A.~Bhattacharjee, R.~Moraffah, J.~Garland, and H.~Liu, ``Towards llm-guided causal explainability for black-box text classifiers,'' in \emph{AAAI 2024 Workshop on Responsible Language Models, Vancouver, BC, Canada}, 2024.

\bibitem{gardner2020evaluating}
M.~Gardner, Y.~Artzi, V.~Basmov, J.~Berant, B.~Bogin, S.~Chen, P.~Dasigi, D.~Dua, Y.~Elazar, A.~Gottumukkala \emph{et~al.}, ``Evaluating models’ local decision boundaries via contrast sets,'' in \emph{Findings of the ACL: EMNLP 2020}, 2020, pp. 1307--1323.

\bibitem{qin2019counterfactual}
L.~Qin, A.~Bosselut, A.~Holtzman, C.~Bhagavatula, E.~Clark, and Y.~Choi, ``Counterfactual story reasoning and generation,'' in \emph{Proceedings of EMNLP-IJCNLP}, 2019, pp. 5043--5053.

\bibitem{bubeck2023sparks}
S.~Bubeck, V.~Chandrasekaran, R.~Eldan, J.~Gehrke, E.~Horvitz, E.~Kamar, P.~Lee, Y.~T. Lee, Y.~Li, S.~Lundberg \emph{et~al.}, ``Sparks of artificial general intelligence: Early experiments with gpt-4,'' \emph{arXiv preprint arXiv:2303.12712}, 2023.

\bibitem{madaan2022plug}
N.~Madaan, S.~Bedathur, and D.~Saha, ``Plug and play counterfactual text generation for model robustness,'' \emph{arXiv preprint arXiv:2206.10429}, 2022.

\bibitem{robeer2021generating}
M.~Robeer, F.~Bex, and A.~Feelders, ``Generating realistic natural language counterfactuals,'' in \emph{Findings of the Association for Computational Linguistics: EMNLP 2021}, 2021, pp. 3611--3625.

\bibitem{li2024prompting}
Y.~Li, M.~Xu, X.~Miao, S.~Zhou, and T.~Qian, ``Prompting large language models for counterfactual generation: An empirical study,'' in \emph{Proceedings of the 2024 Joint International Conference on Computational Linguistics, Language Resources and Evaluation (LREC-COLING 2024)}, 2024, pp. 13\,201--13\,221.

\bibitem{radford2019language}
A.~Radford, J.~Wu, R.~Child, D.~Luan, D.~Amodei, I.~Sutskever \emph{et~al.}, ``Language models are unsupervised multitask learners,'' \emph{OpenAI blog}, vol.~1, no.~8, p.~9, 2019.

\bibitem{brown2020language}
T.~Brown, B.~Mann, N.~Ryder, M.~Subbiah, J.~D. Kaplan, P.~Dhariwal, A.~Neelakantan, P.~Shyam, G.~Sastry, A.~Askell \emph{et~al.}, ``Language models are few-shot learners,'' \emph{Advances in neural information processing systems}, vol.~33, pp. 1877--1901, 2020.

\bibitem{openai2023gpt}
OpenAI, ``Gpt-4 technical report,'' \emph{arXiv}, pp. 2303--08\,774, 2023.

\bibitem{touvron2023llama}
H.~Touvron, T.~Lavril, G.~Izacard, X.~Martinet, M.-A. Lachaux, T.~Lacroix, B.~Rozi{\`e}re, N.~Goyal, E.~Hambro, F.~Azhar \emph{et~al.}, ``Llama: Open and efficient foundation language models,'' \emph{arXiv preprint arXiv:2302.13971}, 2023.

\bibitem{touvron2023llama2}
H.~Touvron, L.~Martin, K.~Stone, P.~Albert, A.~Almahairi, Y.~Babaei, N.~Bashlykov, S.~Batra, P.~Bhargava, S.~Bhosale \emph{et~al.}, ``Llama 2: Open foundation and fine-tuned chat models,'' \emph{arXiv preprint arXiv:2307.09288}, 2023.

\bibitem{dubey2024llama}
A.~Dubey, A.~Jauhri, A.~Pandey, A.~Kadian, A.~Al-Dahle, A.~Letman, A.~Mathur, A.~Schelten, A.~Yang, A.~Fan \emph{et~al.}, ``The llama 3 herd of models,'' \emph{arXiv preprint arXiv:2407.21783}, 2024.

\bibitem{team2023gemini}
G.~Team, R.~Anil, S.~Borgeaud, Y.~Wu, J.-B. Alayrac, J.~Yu, R.~Soricut, J.~Schalkwyk, A.~M. Dai, A.~Hauth \emph{et~al.}, ``Gemini: a family of highly capable multimodal models,'' \emph{arXiv preprint arXiv:2312.11805}, 2023.

\bibitem{penedo2023refinedweb}
G.~Penedo, Q.~Malartic, D.~Hesslow, R.~Cojocaru, A.~Cappelli, H.~Alobeidli, B.~Pannier, E.~Almazrouei, and J.~Launay, ``The refinedweb dataset for falcon llm: Outperforming curated corpora with web data, and web data only,'' \emph{CoRR}, 2023.

\bibitem{gao2020pile}
L.~Gao, S.~Biderman, S.~Black, L.~Golding, T.~Hoppe, C.~Foster, J.~Phang, H.~He, A.~Thite, N.~Nabeshima \emph{et~al.}, ``The pile: An 800gb dataset of diverse text for language modeling,'' \emph{arXiv preprint arXiv:2101.00027}, 2020.

\bibitem{ouyang2022training}
L.~Ouyang, J.~Wu, X.~Jiang, D.~Almeida, C.~Wainwright, P.~Mishkin, C.~Zhang, S.~Agarwal, K.~Slama, A.~Ray \emph{et~al.}, ``Training language models to follow instructions with human feedback,'' \emph{Advances in Neural Information Processing Systems}, vol.~35, pp. 27\,730--27\,744, 2022.

\bibitem{chung2024scaling}
H.~W. Chung, L.~Hou, S.~Longpre, B.~Zoph, Y.~Tay, W.~Fedus, Y.~Li, X.~Wang, M.~Dehghani, S.~Brahma \emph{et~al.}, ``Scaling instruction-finetuned language models,'' \emph{JMLR}, vol.~25, no.~70, pp. 1--53, 2024.

\bibitem{christiano2017deep}
P.~F. Christiano, J.~Leike, T.~Brown, M.~Martic, S.~Legg, and D.~Amodei, ``Deep reinforcement learning from human preferences,'' \emph{Advances in neural information processing systems}, vol.~30, 2017.

\bibitem{dong2024survey}
Q.~Dong, L.~Li, D.~Dai, C.~Zheng, J.~Ma, R.~Li, H.~Xia, J.~Xu, Z.~Wu, B.~Chang \emph{et~al.}, ``A survey on in-context learning,'' in \emph{Proceedings of the 2024 Conference on Empirical Methods in Natural Language Processing}, 2024, pp. 1107--1128.

\bibitem{he2023annollm}
X.~He, Z.~Lin, Y.~Gong, A.~Jin, H.~Zhang, C.~Lin, J.~Jiao, S.~M. Yiu, N.~Duan, W.~Chen \emph{et~al.}, ``Annollm: Making large language models to be better crowdsourced annotators,'' \emph{arXiv preprint arXiv:2303.16854}, 2023.

\bibitem{bansal2023large}
P.~Bansal and A.~Sharma, ``Large language models as annotators: Enhancing generalization of nlp models at minimal cost,'' \emph{arXiv preprint arXiv:2306.15766}, 2023.

\bibitem{tan2024large}
Z.~Tan, A.~Beigi, S.~Wang, R.~Guo, A.~Bhattacharjee, B.~Jiang, M.~Karami, J.~Li, L.~Cheng, and H.~Liu, ``Large language models for data annotation: A survey,'' \emph{arXiv preprint arXiv:2402.13446}, 2024.

\bibitem{sun2023text}
X.~Sun, X.~Li, J.~Li, F.~Wu, S.~Guo, T.~Zhang, and G.~Wang, ``Text classification via large language models,'' \emph{arXiv preprint arXiv:2305.08377}, 2023.

\bibitem{bhattacharjee2024fighting}
A.~Bhattacharjee and H.~Liu, ``Fighting fire with fire: can chatgpt detect ai-generated text?'' \emph{ACM SIGKDD Explorations Newsletter}, vol.~25, no.~2, pp. 14--21, 2024.

\bibitem{nguyen2024llms}
V.~B. Nguyen, P.~Youssef, J.~Schl{\"o}tterer, and C.~Seifert, ``Llms for generating and evaluating counterfactuals: A comprehensive study,'' \emph{arXiv preprint arXiv:2405.00722}, 2024.

\bibitem{chen2022disco}
Z.~Chen, Q.~Gao, A.~Bosselut, A.~Sabharwal, and K.~Richardson, ``Disco: Distilling counterfactuals with large language models,'' \emph{arXiv preprint arXiv:2212.10534}, 2022.

\bibitem{gilardi2023chatgpt}
F.~Gilardi, M.~Alizadeh, and M.~Kubli, ``Chatgpt outperforms crowd workers for text-annotation tasks,'' \emph{Proceedings of the National Academy of Sciences}, vol. 120, no.~30, p. e2305016120, 2023.

\bibitem{maas-EtAl:2011:ACL-HLT2011}
\BIBentryALTinterwordspacing
A.~L. Maas, R.~E. Daly, P.~T. Pham, D.~Huang, A.~Y. Ng, and C.~Potts, ``Learning word vectors for sentiment analysis,'' in \emph{Proceedings of the 49th Annual Meeting of the Association for Computational Linguistics: Human Language Technologies}.\hskip 1em plus 0.5em minus 0.4em\relax Portland, Oregon, USA: Association for Computational Linguistics, June 2011, pp. 142--150. [Online]. Available: \url{http://www.aclweb.org/anthology/P11-1015}
\BIBentrySTDinterwordspacing

\bibitem{maccartney2008modeling}
B.~MacCartney and C.~D. Manning, ``Modeling semantic containment and exclusion in natural language inference,'' in \emph{Proceedings of the 22nd International Conference on Computational Linguistics (Coling 2008)}, 2008, pp. 521--528.

\bibitem{bowman2015large}
S.~Bowman, G.~Angeli, C.~Potts, and C.~D. Manning, ``A large annotated corpus for learning natural language inference,'' in \emph{Proceedings of EMNLP}, 2015, pp. 632--642.

\bibitem{dettmers2023qlora}
T.~Dettmers, A.~Pagnoni, A.~Holtzman, and L.~Zettlemoyer, ``Qlora: Efficient finetuning of quantized llms,'' \emph{arXiv preprint arXiv:2305.14314}, 2023.

\bibitem{ribeiro2016should}
M.~T. Ribeiro, S.~Singh, and C.~Guestrin, ``" why should i trust you?" explaining the predictions of any classifier,'' in \emph{Proceedings of the 22nd ACM SIGKDD international conference on knowledge discovery and data mining}, 2016, pp. 1135--1144.

\bibitem{atanasova2020diagnostic}
P.~Atanasova, J.~G. Simonsen, C.~Lioma, and I.~Augenstein, ``A diagnostic study of explainability techniques for text classification,'' in \emph{Proceedings of EMNLP}, 2020, pp. 3256--3274.

\bibitem{camburu2018snli}
O.-M. Camburu, T.~Rockt{\"a}schel, T.~Lukasiewicz, and P.~Blunsom, ``e-snli: Natural language inference with natural language explanations,'' \emph{Advances in Neural Information Processing Systems}, vol.~31, 2018.

\bibitem{vaswani2017attention}
A.~Vaswani, N.~Shazeer, N.~Parmar, J.~Uszkoreit, L.~Jones, A.~N. Gomez, {\L}.~Kaiser, and I.~Polosukhin, ``Attention is all you need,'' \emph{Advances in neural information processing systems}, vol.~30, 2017.

\bibitem{ali2023explainable}
S.~Ali, T.~Abuhmed, S.~El-Sappagh, K.~Muhammad, J.~M. Alonso-Moral, R.~Confalonieri, R.~Guidotti, J.~Del~Ser, N.~D{\'\i}az-Rodr{\'\i}guez, and F.~Herrera, ``Explainable artificial intelligence (xai): What we know and what is left to attain trustworthy artificial intelligence,'' \emph{Information Fusion}, vol.~99, p. 101805, 2023.

\bibitem{Kaushik2020Learning}
\BIBentryALTinterwordspacing
D.~Kaushik, E.~Hovy, and Z.~Lipton, ``Learning the difference that makes a difference with counterfactually-augmented data,'' in \emph{ICLR}, 2020. [Online]. Available: \url{https://openreview.net/forum?id=Sklgs0NFvr}
\BIBentrySTDinterwordspacing

\bibitem{khashabi2020more}
D.~Khashabi, T.~Khot, and A.~Sabharwal, ``More bang for your buck: Natural perturbation for robust question answering,'' in \emph{Proceedings of EMNLP}, 2020, pp. 163--170.

\bibitem{sanh2019distilbert}
V.~Sanh, L.~Debut, J.~Chaumond, and T.~Wolf, ``Distilbert, a distilled version of bert: smaller, faster, cheaper and lighter,'' \emph{arXiv preprint arXiv:1910.01108}, 2019.

\bibitem{wei2022chain}
J.~Wei, X.~Wang, D.~Schuurmans, M.~Bosma, F.~Xia, E.~Chi, Q.~V. Le, D.~Zhou \emph{et~al.}, ``Chain-of-thought prompting elicits reasoning in large language models,'' \emph{Advances in neural information processing systems}, vol.~35, pp. 24\,824--24\,837, 2022.

\bibitem{cer2018universal}
D.~Cer, Y.~Yang, S.-y. Kong, N.~Hua, N.~Limtiaco, R.~S. John, N.~Constant, M.~Guajardo-Cespedes, S.~Yuan, C.~Tar \emph{et~al.}, ``Universal sentence encoder,'' \emph{arXiv preprint arXiv:1803.11175}, 2018.

\bibitem{levenshtein1966binary}
V.~I. Levenshtein \emph{et~al.}, ``Binary codes capable of correcting deletions, insertions, and reversals,'' in \emph{Soviet physics doklady}, vol.~10, no.~8.\hskip 1em plus 0.5em minus 0.4em\relax Soviet Union, 1966, pp. 707--710.

\bibitem{garg2020bae}
S.~Garg and G.~Ramakrishnan, ``Bae: Bert-based adversarial examples for text classification,'' in \emph{Proceedings of EMNLP}, 2020, pp. 6174--6181.

\bibitem{ribeiro2020beyond}
M.~T. Ribeiro, T.~Wu, C.~Guestrin, and S.~Singh, ``Beyond accuracy: Behavioral testing of nlp models with checklist,'' in \emph{Proceedings of the 58th Annual Meeting of the ACL}, 2020, pp. 4902--4912.

\bibitem{rae2021scaling}
J.~W. Rae, S.~Borgeaud, T.~Cai, K.~Millican, J.~Hoffmann, F.~Song, J.~Aslanides, S.~Henderson, R.~Ring, S.~Young \emph{et~al.}, ``Scaling language models: Methods, analysis \& insights from training gopher,'' \emph{arXiv preprint arXiv:2112.11446}, 2021.

\bibitem{valmeekam2022large}
K.~Valmeekam, A.~Olmo, S.~Sreedharan, and S.~Kambhampati, ``Large language models still can't plan (a benchmark for llms on planning and reasoning about change),'' in \emph{NeurIPS 2022 FMDM Workshop}.

\end{thebibliography}

\end{document}